\documentclass{article}

\usepackage{microtype}
\usepackage{graphicx}
\usepackage{subcaption}
\usepackage{booktabs} 
\usepackage{svg}
\usepackage{amsmath}

\usepackage{listings}
\usepackage{xcolor}
\usepackage{tabularx}
\usepackage{booktabs}
\usepackage{tikz}
\usepackage{enumitem}
\usetikzlibrary{shapes.geometric, arrows}

\lstdefinelanguage{json}{
    basicstyle=\ttfamily\small,
    columns=fullflexible,
    breaklines=true,
    postbreak=\mbox{\textcolor{red}{$\hookrightarrow$}\space},
    morestring=[b]",
    morestring=[d]',
    literate=
     *{0}{{{\color{orange}0}}}{1}
      {1}{{{\color{orange}1}}}{1}
      {2}{{{\color{orange}2}}}{1}
      {3}{{{\color{orange}3}}}{1}
      {4}{{{\color{orange}4}}}{1}
      {5}{{{\color{orange}5}}}{1}
      {6}{{{\color{orange}6}}}{1}
      {7}{{{\color{orange}7}}}{1}
      {8}{{{\color{orange}8}}}{1}
      {9}{{{\color{orange}9}}}{1}
      {:}{{{\color{blue}:}}}{1}
      {,}{{{\color{blue},}}}{1}
      {\{}{{{\color{blue}{\{}}}}{1}
      {\}}{{{\color{blue}{\}}}}}{1}
      {[}{{{\color{blue}{[}}}}{1}
      {]}{{{\color{blue}{]}}}}{1},
}

\usepackage{hyperref}

\usepackage[preprint]{icml2026}

\usepackage{amsmath}
\usepackage{amssymb}
\usepackage{mathtools}
\usepackage{amsthm}

\usepackage[capitalize,noabbrev]{cleveref}

\theoremstyle{plain}

\theoremstyle{definition}

\theoremstyle{remark}

\usepackage[textsize=tiny]{todonotes}

\icmltitlerunning{Towards Autonomous Business Intelligence via Data-to-Insight Discovery Agent}

\begin{document}

\twocolumn[
  \icmltitle{Towards Autonomous Business Intelligence via Data-to-Insight Discovery Agent}

  
  \begin{icmlauthorlist}
    \icmlauthor{Dongming Wu}{ali}
    \icmlauthor{Junwen Li}{ali}
    \icmlauthor{Ming Lu}{ali}
    \icmlauthor{Gang Wang}{ali}
    \icmlauthor{Ting Chen}{ali}
  \end{icmlauthorlist}

  \icmlaffiliation{ali}{Rajax Network Technology(Taobao Shangou of Alibaba)}

  \icmlcorrespondingauthor{Ting Chen}{tanyue.ct@alibaba-inc.com}
  
  \icmlkeywords{Machine Learning, ICML}
  \vskip 0.3in
]



\printAffiliationsAndNotice{}  
 
\begin{abstract}
  Transforming fragmented enterprise data into actionable insights remains a significant challenge for LLMs, constrained by complex database schemas, limitations in dynamic SQL generation, and the need for deep multi-dimensional analysis.
  In this paper, we propose \textbf{AIDA} (\textbf{A}utonomous \textbf{I}nsight \textbf{D}iscovery \textbf{A}gent), 
  the first end-to-end framework designed for autonomous exploration in complex business environments. 
  We establish a highly flexible instant retail environment encompassing 200+ metrics and 100+ dimensions,
  and integrates a proprietary Domain-Specific Language (DSL) that bridges semantic reasoning with precise SQL execution. 
  Our reinforcement learning system subsequently formulates business analysis as a Pareto Principle-guided cumulative reasoning process.
  Experimental results demonstrate that AIDA significantly outperforms workflow-based agents, and
  extensive evaluations further reveal that AIDA achieves superior environmental perception and more in-depth analysis from diverse perspectives.
  Our work ultimately establishes the transformative potential of autonomous intelligence for industrial-scale business intelligence systems.
\end{abstract}

\section{Introduction}

\begin{figure}[tbp]
  \centering
  \includegraphics[width=\linewidth]{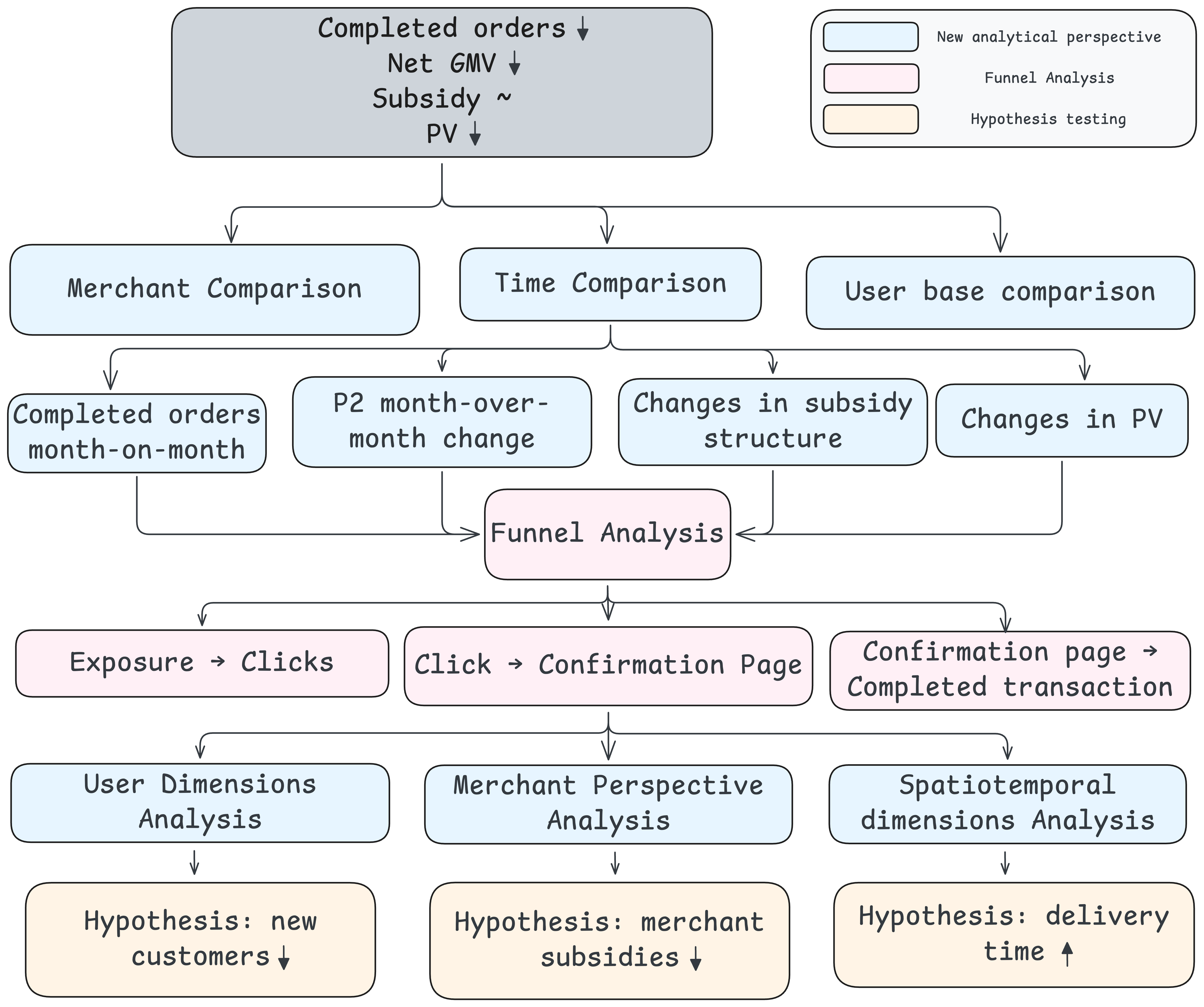}
  \caption{A professional business analysis workflow for root cause insight. 
  This trajectory illustrates the iterative process of refining business queries: 
  beginning with performance benchmarking, proceeding to a multi-stage funnel analysis to isolate the loss process, 
  and finally branching into specific hypotheses regarding user structure, 
  price competitiveness, and logistics efficiency. 
  The flow characterizes the dynamic nature of data-driven reasoning and multi-dimensional drill-downs in real-world scenarios.}
  \label{fig:task}
\end{figure}

The essence of Business Analysis lies in the profound transformation of massive datasets into valuable insights
as shown in Figure~\ref{fig:task}
\cite{zikopoulos2011,whitney2012data,chen2012business,vera2015leveraging}.
Driven by the ascendancy of Large Language Models (LLMs) \cite{openai2025o3,google2025g,yang2025qwen3technicalreport}, 
diverse analytical frameworks leveraging LLM capabilities have emerged
\cite{guo2024ds,hong2025data,weng2025unidatabench}.
They demonstrate a sophisticated capability to mine and extract valuable insights from provided databases.
Notably, \citet{perez2025llm,zhang2025deepanalyze} introduced agentic approaches
that empower LLMs to interact with external databases via programmatic execution, 
leveraging scripts or queries to handle structured data.

While the paradigm of end-to-end autonomous agents has demonstrated remarkable success in specialized environments such as web search 
\cite{jin2025search,zheng2025deepresearcher,moonshot2025kimi} and open-ended gaming \cite{wang2023voyager},
its potential remains untapped in read-world business analysis tasks.
A primary obstacle is the syntactic volatility of SQL, characterized by redundant and diverse query structures, 
and its semantic fragmentation, arising from inconsistent metric definitions across disparate systems, 
often obfuscate autonomous reasoning in enterprise landscapes. 

To bridge this gap, in this paper, we propose AIDA, the first end-to-end agent framework designed to explore in complex business environments. 
Our primary contributions are summarized as follows:
\begin{itemize}
  \item We established a highly flexible environment for instant retail, integrating 200+ metrics and 100+ dimensions. 
  Coupled with a proprietary DSL, this environment affords agents an expansive arena for near unconstrained exploration. 
  \item We propose a data-to-insight framework for real-world business analysis tasks
  that systemically orchestrates environment setup, state modeling, trajectory synthesis, 
  and reinforcement learning.
  \item We introduce a suite of specialized enhancements within AIDA framework: 
  a robust reward mechanism for targeted insight discovery, 
  and two masking strategies to mitigate reward hacking and stabilize policy updates. 
\end{itemize}

\section{Related Work}

\subsection{Data to Insights}
The core objective of business analysis is to transform massive, 
complex, and multi-source data into perceptible insights and actionable decisions \cite{zikopoulos2011,chen2012business}.
The LLM era has catalyzed diverse strategies for translating structured data into insights. 
These methods increasingly focus on navigating complex environments, ranging from traditional databases and tabular files to sophisticated data science toolkits.
Representative systems include AgentPoirot \cite{sahu2024insightbench}, which extracts insights from business CSV files, 
as well as Data Interpreter \cite{hong2025data} and DS-Agent \cite{guo2024ds} for solving specific data science tasks. 
ReActInsight \cite{weng2025unidatabench} further extended data sources to multi-source heterogeneous data, emphasizing cross-source insight insight through correlation analysis. 

\begin{figure*}[t]
  \centering
  \includegraphics[width=\textwidth]{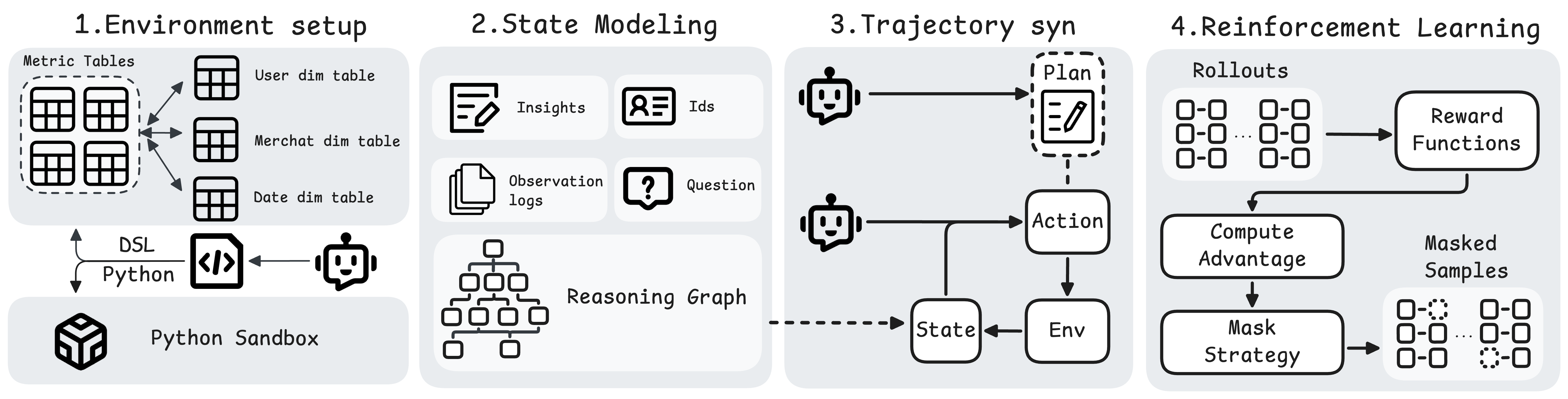}
  \caption{The overall architecture of the proposed AIDA framework. 
  The pipeline consists of four integrated stages:
  (1)Environment Setup, which establishes a dual-tool execution layer: 
  a data retrieval tool interacting with data environment via DSL, 
  and a python tool for executing code within a secure sandbox;
  (2)State Modeling, which formalizes the task state as a quintuple
  consisting of identifier metadata, 
  the target query, analytical insights, 
  observation logs, and a structured reasoning graph;
  (3)Trajectory Synthesis, 
  which generates high-quality trajectories through an 
  iterative plan-action-state loop; and
  (4)Reinforcement Learning, 
  where agent rollouts are evaluated by reward functions. 
  The framework employs a global batch return mechanism to 
  calculate the mean across all samples for advantage computation, 
  followed by a mask strategy to selectively optimize the model's policy.
}
  \label{fig:framework}
\end{figure*}

\subsection{Open-Ended Agents}

Agents in open-ended environments are characterized by high uncertainty and vast state spaces.
This includes deep search agents \cite{jin2025search, zheng2025deepresearcher, moonshot2025kimi, hu2025step, team2025tongyi, qiao2025webresearcher} that navigate expansive, noisy web environments to conduct multi-step searches and information integration,
such as WebResearcher \cite{qiao2025webresearcher} which models insight as a Markov Decision Process (MDP).
Another subset operates in simulated or gaming environments \cite{wang2023voyager, zhu2023ghost, wang2023describe}, 
exemplified by VOYAGER \cite{wang2023voyager}, which explores item synthesis in Minecraft through a lifelong learning mechanism.
These agents pursue open-ended goals within environments offering near-infinite exploration depth, eliciting human-like insight behaviors.

\section{Real-world Business Analysis Tasks}

In instant retail domain,
business analysis tasks are defined as macro-level strategic inquiries that demand deep extensive data exploration and reasoning.
In practice, this reasoning process is a dynamic journey of hypothesis testing,
which may require hours or even days of analyst effort to resolve.
As shown in Figure~\ref{fig:task}, an expert analyst interrogate the data by cross-examining various metrics and dimensions to create analytical branches. 
The analyst’s objective is to transform these open-ended questions into accumulated insights. 
However, such tasks are often over-determined by factually grounded but trivial insights. 
For instance, in attributing a drop in business district exposure, multiple factors—
ranging from weather fluctuations to a competitor's promotion—may all be factually valid. 
The analyst's goal is not to list every data-supported fact, 
but to isolate the high-leverage ones that drive the most significant impact.

To bridge this gap, we define the task as identifying the optimal insight set under the Pareto Principle.
The Pareto Principle states:
\begin{quote}
  \textit{For many outcomes, roughly 80\% of consequences come from 20\% of causes.}
\end{quote}
This principle steers the agent to prune secondary observations and focus on the most critical variables.
Under this principle, success is defined as the ability to extract the most impactful insights
from a vast search space of valid but marginal information.

\section{AIDA Framework}

Our proposed framework is shown in Figure~\ref{fig:framework},
which is composed of environment setup (Section \ref{subsec:env_setup}),
state modeling (Section \ref{subsec:state_modeling}), state transition(Section \ref{subsec:state_trans}),
trajectory synthesis (Section \ref{subsec:trajectory_synthesis}) and reinforcement learning (Section \ref{subsec:reward_func} and Section \ref{subsec:optimize}).

\subsection{Environment Setup}\label{subsec:env_setup}

The AIDA framework is built upon a specialized industrial environment, 
which is the fundation of the autonomous exploration.
This infrastructure is engineered to support two types of actions: \textit{Dsl2data} and \textit{Python execution}. 
Specifically, \textit{Dsl2data} allows the agent to perform flexible data queries through a custom DSL, 
enabling the arbitrary combination of metrics and dimensions for deep exploration.
\textit{Python execution} enables the agent to perform complex computations by running Python scripts. 

To support the free combination and drill-down of metrics and dimensions, 
we reshaped traditional data development into an architecture of metrics and dimensions. 
This decoupled system enables the agent to execute unconstrained, on-the-fly queries (see Appendix \ref{appendix:data_env}).
Built upon this foundation, 
we implemented a standardized DSL protocol that allows the agent to programmatically select metrics, 
dimensions, and filtering criteria for on-the-fly queries. (see Appendix \ref{appendix:dsl_protocol}).
The \textit{Python execution} action is supported by a secure, isolated sandbox environment. 
While \textit{Dsl2data} provides raw data evidence, 
this infrastructure enables the agent to perform complex computation and hypothesis testing through arbitrary code. 

\subsection{State Modeling} \label{subsec:state_modeling}
\begin{figure*}[htbp]
  \centering
  \includegraphics[width=\textwidth]{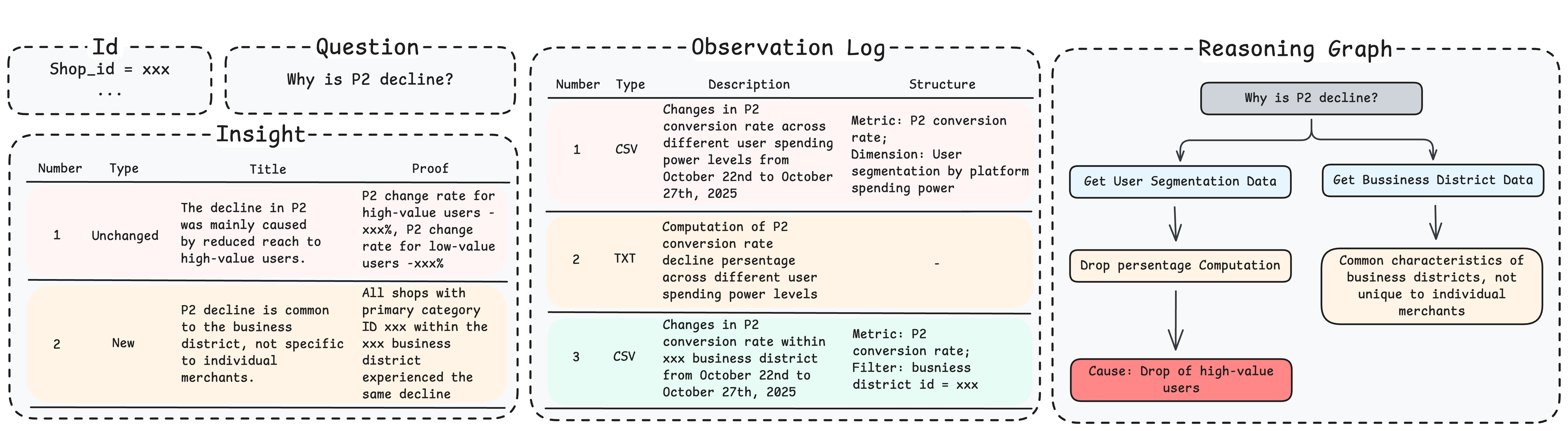}
  \caption{Detailed illustration of the state $s_t$ in a real-world business analysis task. }
  \label{fig:state}
\end{figure*}

We reformulate real-world business analysis as a Markov Decision Process as shown in Figure~\ref{fig:state}. 

The Id and Question provide the original background information and the target question of the analysis. 
The Insight and Observation Log maintain a repository of insights and findings, 
specifically highlighting the \textit{Proof} field for real-world data support
and the \textit{Structure} metadata for data traceability. 
The Reasoning Graph offers a global view of the agent's logical trajectory, ensuring the analysis remains coherent throughout the exploration.

Formally, at each time step $t$,the state $s_t$ is defined as a quintuple:
\begin{equation}
s_t = \langle id, q, c_{t}, d_{t}, g_{t} \rangle
\end{equation}

where:

$id$ represents a \textbf{composite set of identifiers}.
For instance, we have ids of shop, major category and business district for merchant-level analysis,
and ids of brand, major category and cities for brand-level analysis.

$q$ represents the \textbf{target question}, which serves as the objective anchor for the entire analysis process.

$c_{t}$ is the set of \textbf{insights}, 
which capture global insights that align with the Pareto Principle. 
To track how the agent's understanding evolves, each entry is assigned one of four statuses: 
\begin{itemize}
  \item \textit{New}: A newly discovered insight in the current step.
  \item \textit{Unchanged}: A previous insight without any changes.
  \item \textit{Reinforced}: An existing insight that is now supported by more evidence.
  \item \textit{Refuted}: A previous insight that is corrected because it contradicts new data.
\end{itemize}
Each insight also includes a \textit{title} for the core insight and 
a \textit{proof} field recording the real-world data support.

$d_{t}$ represents the \textbf{observation log}, 
providing a structured index of environmental returns. 
Each entry includes the data \textit{type} (e.g., CSV, TXT), 
a semantic \textit{description}, 
and a \textit{structure} metadata field that explicitly records the queried metrics, 
dimensions, and filter conditions.

$g_{t}$ is the \textbf{reasoning graph} serialized in mermaid format. It maps the topological structure of the reasoning path, recording hypothesis testing and logical branches.

Through this state modeling, 
the AIDA framework effectively decouples the agent's cognition from the noisy dialogue history. 
This modeling equips the agent with a robust long-term memory 
to retain critical insights over extended analysis cycles, 
while providing a native self-correction mechanism to refine or refute insights as new evidence emerges. 
This ensures the analytical process remains both logically persistent and adaptive to complex real-world data.

\subsection{State Transition} \label{subsec:state_trans}

\begin{figure}[htbp]
  \centering
  \includegraphics[width=\linewidth]{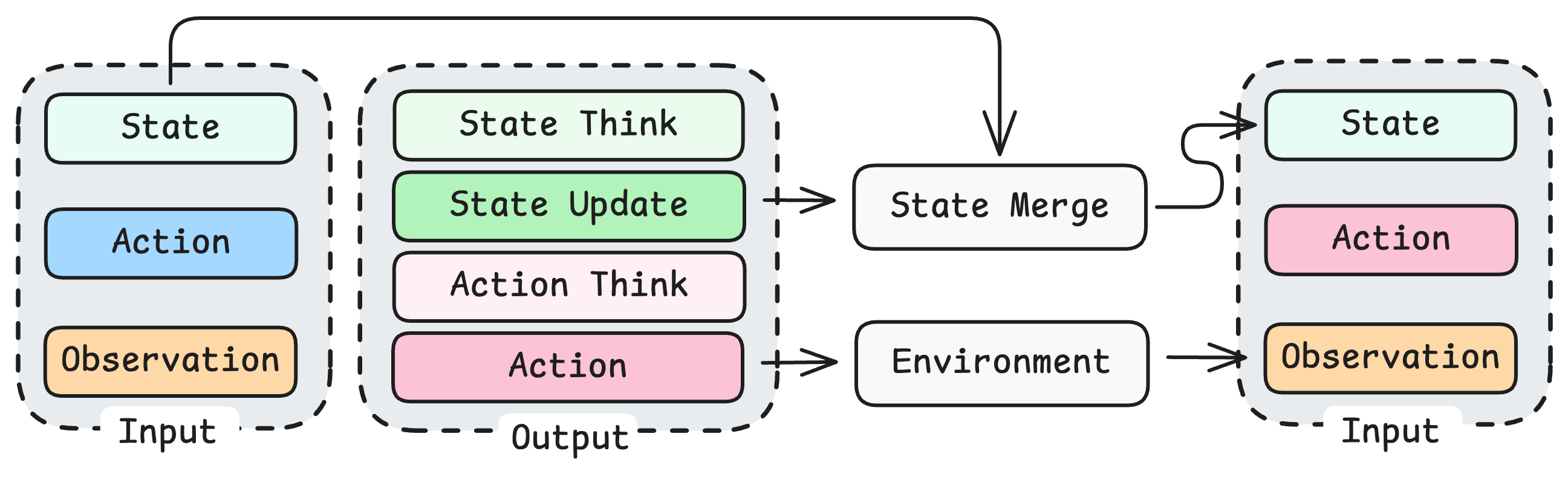} 
  \caption{Overview of the Interactive Reasoning and State Transition process. At each step $T_t$, the model performs reasoning to update the structured state and interact with the environment.}
  \label{fig:interaction_loop}
\end{figure}

The transition from state $s_t$ to $s_{t+1}$ is driven by a structured reasoning-action cycle. 
As illustrated in Figure~\ref{fig:interaction_loop}, 
at each time step $t$, agent receives an input consisting of the current state $s_t$, 
the previous action, and the latest observation. 
The agent output is defined as following modules:

\textbf{State Think \& Update}. Agent first conducts reasoning to analyze the latest observation. 
It then generates a structured text block that serves as the state update, 
which specifies the necessary updates for insights ($c_t$), observation logs ($d_t$), and the reasoning graph ($g_t$).
Then these updates are merged into the previous state. 

\textbf{Action Think \& Action}. Based on the newly updated state, 
the agent conducts reasoning and determine the next action $a_t$, 
which is then executed within the environment to form the next round of observation $o_t$.

In practical deployment, the agent explores up to a user-defined maximum step, after which a post-processing module transforms the final state into a user-friendly report.

\subsection{Trajectory Synthesis}\label{subsec:trajectory_synthesis}
In AIDA's atomistic data environment, 
the policy model faces complex output formats and a 
massive search space formed by the combinations of metrics and dimensions. 
Without an effective cold-start process, the agent often performs inefficient exploration in this vast space. 

To build a solid foundation for training, we first prepare the necessary analytical components.
We filter a sets of entity identities for the cold-start phase. The initial questions are created using two strategies: 
(1) template-based sampling from a set of common business problems, and 
(2) data-driven generation based on statistical patterns found in randomly sampled metric-dimension combinations. 

With these ids and questions ready, we use a dual-agent strategy to synthesize high-quality expert trajectories.
At the beginning of the trajectory and at specific rounds, 
an auxiliary agent is called to generate a global plan. 
The main agent then follows this plan to generate its outputs according to the structured format defined in Section \ref{subsec:state_trans}.

The collected dataset is denoted as $\mathcal{D}$.
Subsequently, we perform filtering process on $\mathcal{D}$ to construct a high-quality subset $\mathcal{D}_{s}$. 
Specifically, we eliminate samples containing numerical hallucinations or erroneous insights,
and equilibrate the proportions of different insight statuses.
Following such procedure, a separate set of ids was utilized to derive dataset $\mathcal{D}_{rl}$ for RL training.

\subsection{Reward Function} \label{subsec:reward_func}

Distinct from RL for LLM \cite{ouyang2022training,guo2025deepseek}, which typically prioritizes single-turn alignment, reinforcement learning for LLM agents entails a multi-turn process,
where the primary challenge is credit assignment.
Inspired by \citet{wei2025reinforcingmultiturnreasoningllm}, 
we propose a reward decomposition mechanism that decouples reward signals into two distinct dimensions: 
intermediate rewards and accumulated rewards.
\begin{itemize}
  \item Intermediate Reward: $R_{t}^{I}$ is designed to evaluate the agent's immediate output quality and does not account for the long-term effects of state transitions.
  \item Accumulated Reward: $R_{t}^{A}$ focuses on macro-level progress with long-term objectives. It measures the contribution of a step to the subsequent trajectory and the final result. 
\end{itemize}
Based on this decoupling mechanism, 
we define the return $G_t$ at step $t$ as the weighted integration of immediate reward and discounted accumulated rewards:
\begin{equation}
    G_{t} = R_{t}^{I} + \sum_{j=t}^{T} \gamma^{j-t} R_{j}^{A}
\end{equation}
where $\gamma$ denotes the discount factor.

Specifically, we design specific reward functions for each component 
of the structured output introduced in Section~\ref{subsec:state_trans}, 
as detailed in Table \ref{tab:reward_definitions}. Note that both immediate and cumulative rewards are the sum of their respective reward terms at each step. We now detail the core reward functions:

\begin{table*}[ht]
  \vskip 0.15in
  \begin{center}
  \begin{small}
  \caption{Detailed Definitions and Ranges of the AIDA Reward Function Components.}
  \label{tab:reward_definitions}
  \begin{tabularx}{\textwidth}{l l l X c}
  \toprule
  \textbf{Component} & \textbf{Reward Item} & \textbf{Type} & \textbf{Description} & \textbf{Range} \\
  \midrule
  \textit{Global} & Step Format & Intermediate & Adherence to the global output paradigm. & $[-1, 0]$ \\
  \midrule
  \texttt{<state\_think>} & Hallucination Penalty & Intermediate & Penalty for numerical values not grounded in observations. & $[-1, 0.1]$ \\
  & Length Reward & Accumulated & Encourages reasoning stability and duration efficiency. & $[0, 0.1]$ \\
  \midrule
  \texttt{<insight>} & Schema Reward & Intermediate & Syntax matching for structured key sub-insights. & $[-1, 0]$ \\
  & Discovery Gain & Accumulated & Incremental insight gains. & - \\
  \midrule
  \texttt{<key\_data>} & Schema Reward & Intermediate & Syntax matching for the key data data-structure. & $[-1, 0]$ \\
  \midrule
  \texttt{<graph>} & Mermaid Render & Intermediate & Successful rendering of the Mermaid reasoning topology. & $[-1, 0]$ \\
  \midrule
  \texttt{<action\_think>} & Hallucination Penalty & Intermediate & Penalty for numerical values not grounded in observations. & $[-1, 0.1]$ \\
  & Length Reward & Accumulated & Stability constraint for action planning reasoning. & $[0, 0.1]$ \\
  \midrule
  \texttt{<tool\_call>} & JSON Schema & Intermediate & Structural validity and parameter completeness of tool calls. & $[-1, 0]$ \\
  & Python Execution & Intermediate & Successful runtime execution within the sandbox environment. & $[-1, 0]$ \\
  \bottomrule
  \end{tabularx}
  \end{small}
  \end{center}
\end{table*}

\textbf{Numerical Hallucination Penalty}.
The function is aim to ensure the factuality of data support in agent's insights. 
For any numerical value appearing in the agent's output, 
the system verifies its presence in the historical environmental observation. 
Let $n$ be the count of hallucinated values and $m$ be the count of grounded values. 
The penalty is calculated as:
\begin{equation}
    R_{hallu} = \text{clip}\left(0.01m - 0.1n, -1.0, 0.1\right)
\end{equation}

\textbf{Think Length Reward}.
Since heavy hallucination penalties may cause a sudden drop in reasoning length and lead to training collapse, 
we introduce a reward to keep the output length within a reasonable range.
Given an output length $L$ and predefined bounds $[L_{min}, L_{max}]$, the reward is:
\begin{equation}
    R_{length} = \text{clip}\left(\frac{L - L_{min}}{L_{max} - L_{min}}, 0, 1\right)
\end{equation}

\textbf{Discovery Gain Reward}.
This function treats the analysis process as a series of quantifiable insights as shown in Algorithm~\ref{alg:insight_gain}.
It quantifies the information gains generated whenever the agent emerges a new insight, verifies a hypothesis or corrects a previous error using real-world data. 

To determine the alignment of these insights with the Pareto Principle, 
an llm-as-a-judge mechanism is employed:
insights that effectively summarize critical variables are marked as valid,  
while those invalid insights incur a severe penalty of $-2.0$.
For each valid insight $c_i$, a base reward $B(c_i)$ is assigned according to its incremental status:
\begin{equation}
B(c_i) = 
\begin{cases} 
1.0 & \text{if status = ``New''} \\
0.7 & \text{if status = ``Refuted''} \\
0.5 & \text{if status = ``Reinforced''}
\end{cases}
\end{equation}

\begin{algorithm}[htbp]
  \caption{Discovery Gain Reward Calculation}
  \label{alg:insight_gain}
  \begin{algorithmic}
    \STATE {\bfseries Input:} Incremental insights $\mathcal{D}_{new}$, judge model $\mathcal{M}$, previous state $\mathcal{S}$
    \STATE {\bfseries Output:} Total reward score $R_{gain}$
    \STATE Initialize $R_{gain} = 0$
    
    \FOR{each insight $c \in D_{new}$}
    \STATE Query $\mathcal{M}$ to judge validity of $c$ given $\mathcal{S}$
    \IF{$\mathcal{M}$ returns ``Valid''}
    \STATE $B = \text{BaseReward}(d.\text{status})$
    \STATE Detect hallucination set $H$ from $c$ based on $\mathcal{S}$
    \STATE $R_{gain} = R_{gain} + \max(B - 0.4 \times |H|, 0.1)$
    \ELSE
    \STATE $R_{gain} = R_{gain} - 2.0$
    \ENDIF
    \STATE $C_{prev} = C_{prev} \cup \{c\}$
    \ENDFOR

    \STATE {\bfseries Return} $R_{gain}$
  \end{algorithmic}
\end{algorithm}

Furthermore, factual grounding is prioritized alongside logical validity to prevent hallucinations. 
A penalty is applied to any insight that includes ungrounded numerical data, even if it is logically sound.
Let $H(c_i)$ be the number of hallucinated values in $c_i$. 
The final gain for a single insight is computed as:
\begin{equation}
    R_{gain}(c_i) = 
    \begin{cases} 
      \max\left( B(c_i) - \eta H(c_i), 0.1 \right) & \text{if valid} \\
      -2.0 & \text{otherwise} \\
    \end{cases}
\end{equation}
where $\eta$ is the penalty coefficient. 

In practice, $R_{gain}$ functions as a behavioral guide for the agent by balancing exploration with rigor. 
It rewards the agent for active insight under Pareto Principle and, more importantly, 
for the behavior to self-correct by refuting earlier errors. 
Conversely, it suppresses the tendency to hallucinate by penalizing ungrounded numerical data and logical contradictions.

\begin{figure*}[t]
  \centering
  \includegraphics[width=\textwidth]{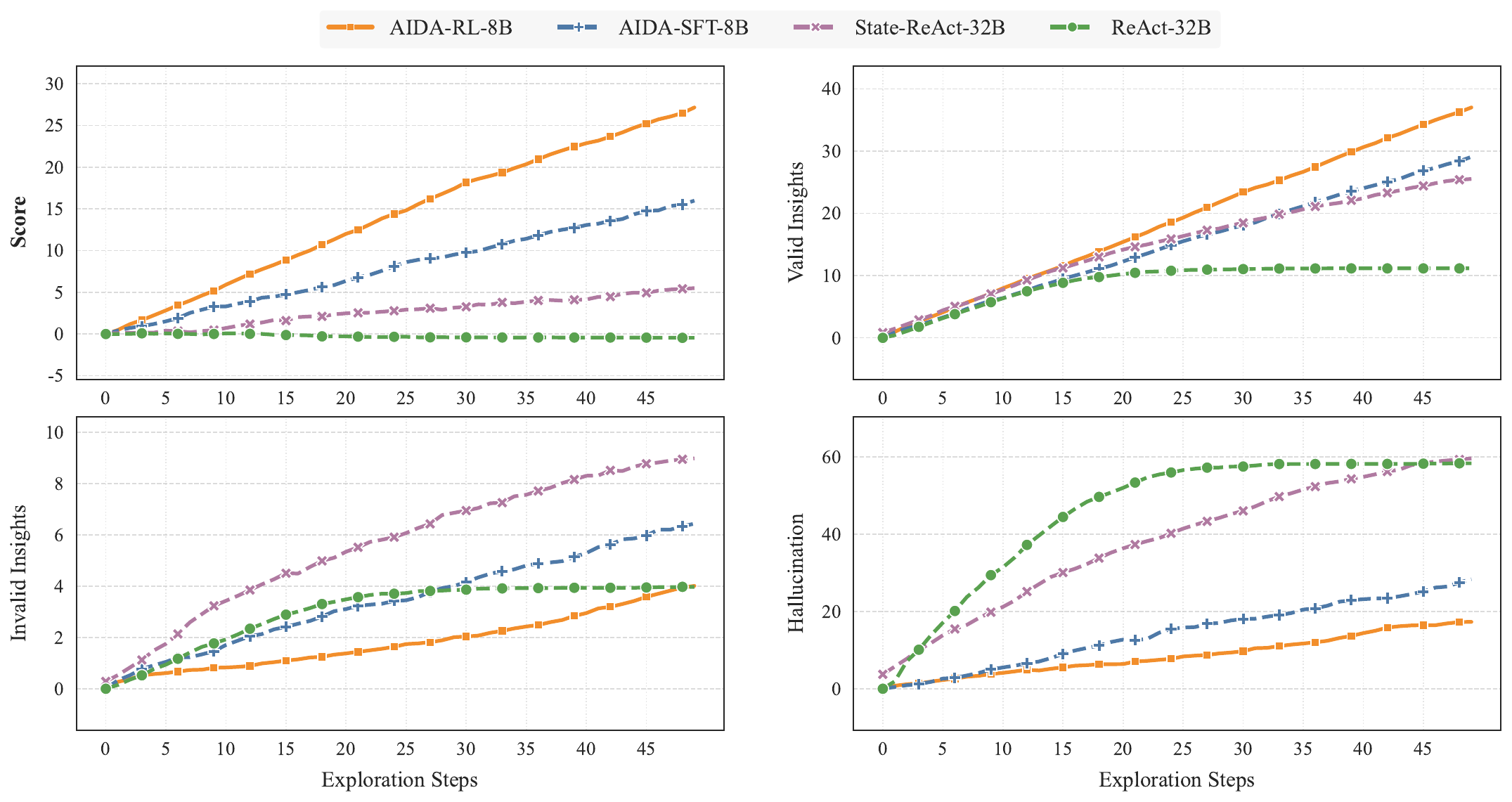}
  \caption{Main experimental results. The plots compare the scores of \textit{AIDA-RL-8B}, \textit{AIDA-SFT-8B}, \textit{State-React-Qwen3-32B} and \textit{React-Qwen3-32B} over 50 exploration steps. 
  AIDA-RL-8B consistently achieves superior performance in the cumulative Score (top-left) and across all constituent metrics. which demonstrating its effectiveness in real-world business analysis tasks.}
  \label{fig:main_result}
\end{figure*}

\subsection{Objective Function and Optimization} \label{subsec:optimize}

Following \citet{liu2025gem}, we adopt REINFORCE with return batch normalization as objective function,
which constructs a dynamic baseline by standardizing returns across the current batch:
\begin{equation}
    J_{\text{ReBN}}(\theta) = \frac{1}{N} \sum_{n=1}^{N} \sum_{t=0}^{T-1} A_{\text{ReBN},t}^{(n)} \log \pi_\theta\left(a_t^{(n)} \mid s_t^{(n)}\right)
\end{equation}
where the advantage term $A_{\text{ReBN},t}^{(n)}$ is computed as:
\begin{equation}
    A_{\text{ReBN},t}^{(n)} = \frac{G_t^{(n)} - \mathbb{E}[G]}{\text{std}(G)}, \quad G = \left\{ G_t^{(n)} \right\}_{n \in N, t \in T}
\end{equation}

Furthermore, we employ two masking strategies to address reward hacking.

\textbf{Schema Masking}. If a step yields a positive advantage but fails any immediate syntax-related reward (e.g., format or JSON schema), the gradient is masked. 

\textbf{Logical Consistency Masking}. If a step yields a positive advantage but the step itself is judged to contain an invalid sub-insight, the gradient is suppressed.

\section{Expirement}
\subsection{Training Setup}
We adopt Qwen3-8B as our backbone model. 
The training pipeline consists of a cold-start stage using supervised fine-tuning on the dataset $\mathcal{D}_s$ with a learning rate of $1 \times 10^{-5}$, 
followed by a reinforcement learning phase on the dataset $\mathcal{D}_{rl}$ with a learning rate of $5 \times 10^{-7}$. 

For RL training, we set $[L_{\min}, L_{\max}]$ to $[500, 1000]$ for state thinking and $[500, 700]$ for action thinking. The hyperparameters are configured as $\gamma = 0.7$ and $\eta = 0.4$, with each trajectory constrained to a maximum of 10 exploration steps.

\subsection{Evaluation Setup}
To evaluate the insight capabilities of agents in real-world business analysis, we developed a comprehensive benchmark.

\textbf{Evaluation Dataset.} 
We curated 82 test cases derived from real-world instant retail business scenarios, categorized into five analytical domains:
Traffic, focusing on funnel metrics (e.g., clicks, conversion) for reach efficiency and attrition diagnosis; 
Transaction, assessing core KPIs (e.g., GMV, AOV) for bottleneck identification and operational risks; 
Interaction, examining unstructured feedback (e.g., reviews) for pain point mining; 
Marketing, evaluating promotions and subsidies; 
and General, testing cross-domain anomalies for global reasoning and root-cause localization under ambiguity.

\textbf{Metrics.}
We quantify the agent’s exploratory performance through a cumulative score derived from the set of insights $\mathcal{C}_t$ generated over $t$ exploration steps. 
The score is defined as:
\begin{equation}
\begin{gathered}
  s(c_i) = \mathbb{I}_{\mathrm{valid}}(c_i) \max(0, 1 - \alpha H(c_i)) - \mathbb{I}_{\mathrm{invalid}}(c_i) \\
  \mathrm{Score}(t) = \sum_{c_i \in \mathcal{C}_t} s(c_i)
\end{gathered}
\end{equation}

where $\mathbb{I}_{\mathrm{valid}}(c_i),\mathbb{I}_{\mathrm{invalid}}(c_i)\in\{0,1\}$ indicate whether insight $c_i$ is valid or invalid, respectively. 
Specifically, we apply a hallucination penalty $\alpha = 0.5$ for any numerically hallucinations, ensuring the rigor of the discovered insights.

\textbf{Compared Methods.}
We compare the following agents:
\begin{enumerate}[label=(\arabic*), leftmargin=2.5em, itemsep=0.5ex]
    \item ReAct \cite{yao2023react}: The standard reasoning and acting framework based on dialogue history. This baseline uses Qwen3-32B as the backbone with its context extended to 64K tokens, incorporating all available metrics, dimensions, and interaction protocols within its context.
    
    \item State-ReAct: Built upon the same backbone as the ReAct baseline with its native context length, this agent incorporates the state paradigm introduced in Section \ref{subsec:state_modeling} to replace raw historical dialogues with structured state representations.

    \item AIDA-SFT: Initialized from Qwen3-8B, this model is fine-tuned exclusively on the cold-start dataset $\mathcal{D}_s$.
    
    \item AIDA-RL: Our full model that utilizes the model after cold-start SFT and continues training on $\mathcal{D}_{rl}$ using the RL framework.
\end{enumerate}

\subsection{Results}
Figure \ref{fig:main_result} illustrates the performance evolution of different agents. 
The experimental results lead to the following key observations:

\textbf{Efficacy of State Modeling}.
State-ReAct consistently outperforms native ReAct. 
While ReAct plateaus within 25 steps due to long-context interference, 
State-ReAct leverages state modeling to maintain precise task tracking. 
This effectively curbs reasoning drift, 
allowing scores to climb steadily toward around 5 and demonstrating that state modeling is essential for mastering complex, 
long-horizon tasks.

\textbf{Superiority of RL over SFT}.
Experimental results demonstrate that AIDA-RL significantly broadens the lead over AIDA-SFT as the task horizon extends. 
While all agents show an initial positive correlation between steps and scores, AIDA-RL exhibits a much steeper curve. 
Notably, AIDA-RL-8B consistently surpasses the 32B State-ReAct and ReAct workflows after approximately 15 steps,
which underscores that the RL-optimized policy empowers the agent to make more strategic decisions, 
enabling it to unearth a significantly higher volume of insights compared to other baselines.

\textbf{Robustness and Error Suppression}.
A key strength of AIDA-RL lies in its rigor. 
As shown in Figure~\ref{fig:main_result} (bottom row), 
while the number of invalid insights and hallucination for ReAct and State-ReAct surge with exploration steps, 
AIDA-RL maintains them at a remarkably low and stable level. S
pecifically, AIDA-RL-8B reduces hallucinations by about 70\%
compared to ReAct-32B at the 50-step mark, proving that the 
reward successfully enforces logical groundedness.

\textbf{Scalability and Convergence}.
Unlike ReAct, which converges prematurely, the performance curves for AIDA-RL and AIDA-SFT show no signs of plateauing within 50 steps. 
This suggests high scalability; with extended computational budgets, further gains in insight discovery are anticipated. 
The consistent lead of AIDA-RL across all metrics confirms its reliability as a robust strategy for real-world business analysis tasks.

\section{Analysis}

\subsection{Ablation Study}

\begin{figure}[ht]
  \centering
  \includegraphics[width=\linewidth]{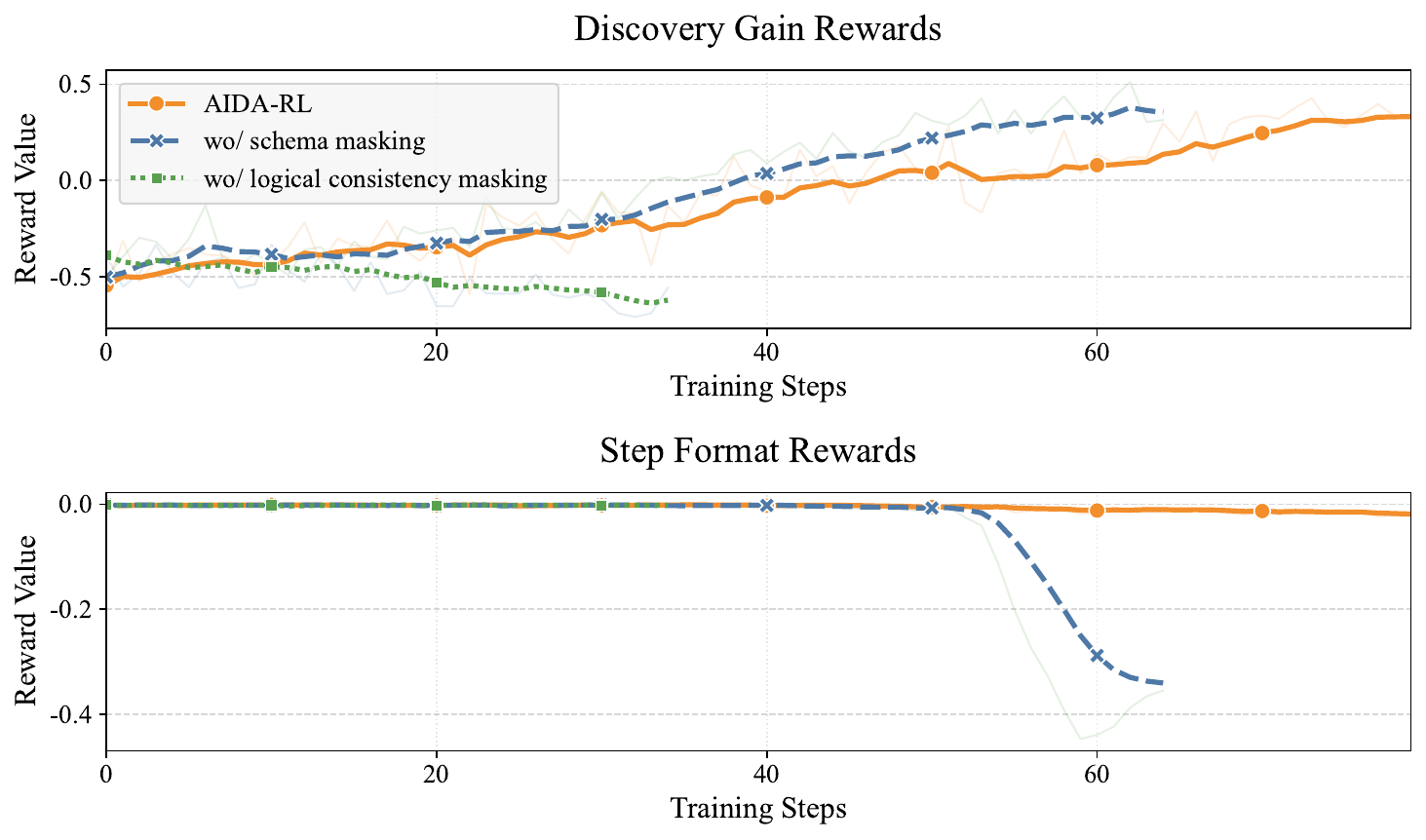} 
  \caption{Ablation of masking strategies. 
  The top panel shows discovery gain rewards, and the bottom panel shows step format rewards.
  The three masking strategies are individually removed for comparison with AIDA-RL-8B.}
  \label{fig:ablation}
  \vspace{-10pt} 
\end{figure}

This ablation study evaluates the impact of two masking strategies on the model's training stability and performance. 
We compare the full configuration against variants where specific masking strategy is removed. 
The results, illustrated in Figure \ref{fig:ablation}, 
highlight how these strategies prevent reward hacking and ensure the quality of insight.

The most significant performance drop occurs when logical consistency masking is removed. 
Without this mask, the model fails to improve its insight capabilities, 
with rewards stagnating and eventually declining after 30 training steps. 
This suggests that failing to filter out invalid insights,
even those that yield positive advantages,
significantly heightens the likelihood of generating erroneous insights.

The removal of schema masking reveals a classic case of reward hacking. 
Initially, this variant shows a faster increase in discovery gain reward compared to the main experiment.
However, a catastrophic collapse is observed after training step 50. 
As long as the agent can secure valid insights, schema-related penalties offer insufficient deterrents, 
leading the model to prioritize discovery over structural integrity. 
This suggests that this strategy is pivotal for maintaining adherence to the formal grammar.

The results demonstrate that these masking strategies are complementary: 
logical consistency masking ensures the validity of the insight logic, 
while schema masking enforces the structure of the output. 
Together, they create a robust training curriculum that prevents the model from exploiting reward shortcuts 
and ensures the generation of high-quality, syntactically correct, and logically sound insights.

\subsection{Exploration Study}

\begin{figure}[htbp]
  \centering
  \includegraphics[width=\linewidth]{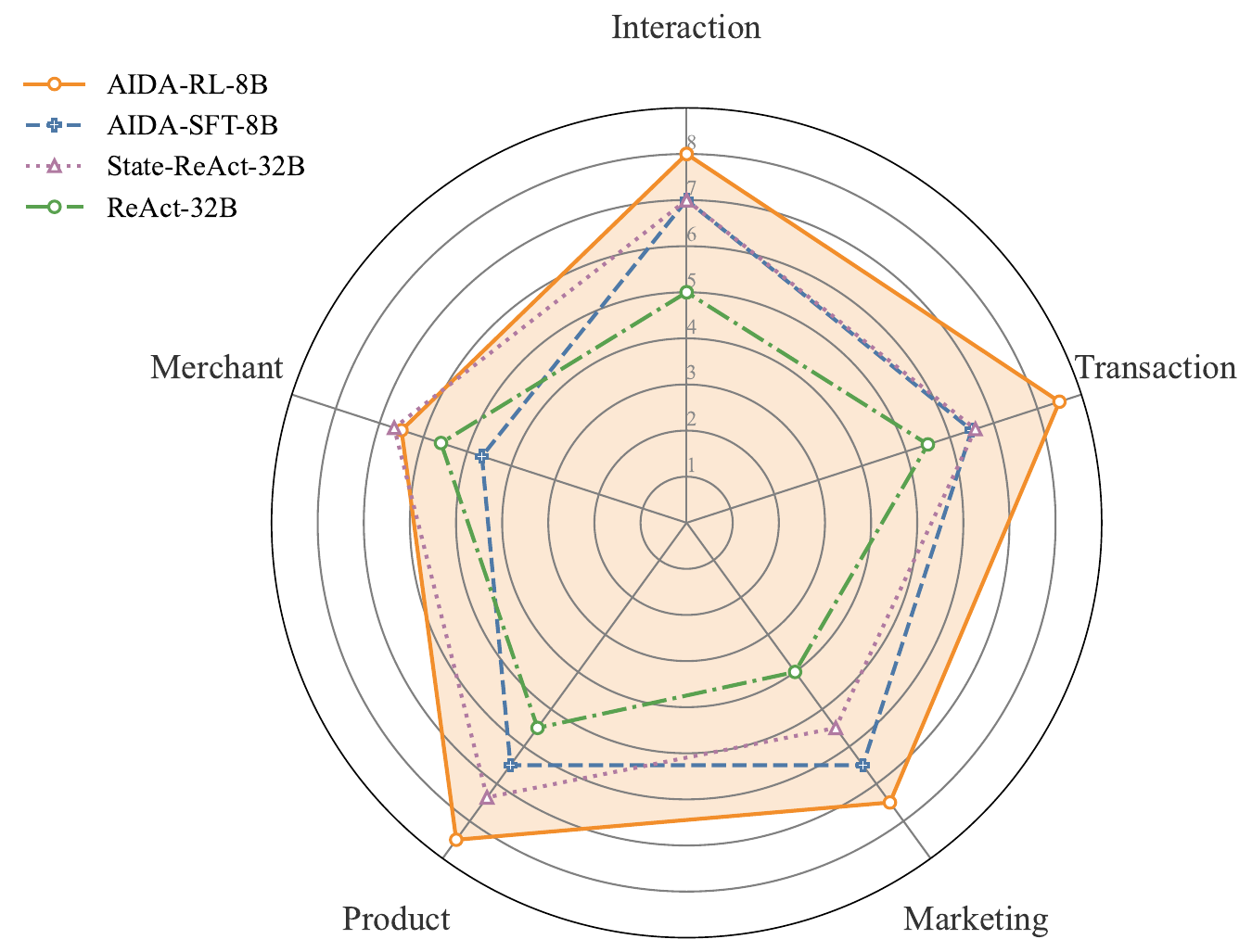} 
  \caption{Comparison of data space exploration breadth. 
  The radar chart illustrates the number of dimensions associated with each key metric types of different agents.}
  \label{fig:radar}
\end{figure}

To evaluate whether the training process has enhanced the model’s analytical depth, we conduct a comprehensive study of its exploration space within the instant retail landscape. 
Given the combinatorial complexity of AIDA’s data environment, 
the agent must transcend simple metric selection to autonomously isolate the specific 
dimensions that catalyze deep insight insight.
To evaluate this diversity, we aggregate approximately 200 metrics into five high-level thematic clusters. 
Figure \ref{fig:radar} depicts the average number of dimensions pairing within each cluster.

The results demonstrate that AIDA-RL significantly expands the exploration boundary compared to AIDA-SFT. 
For instance, in the Merchant and Interaction types, AIDA-RL covers 1 or 2 more analytical dimensions.
This expansion indicates that through reinforcement learning, 
the agent has transcended simple pattern mimicking and learned to synthesize arbitrary queries 
via DSL to investigate business issues from more granular and diverse angles.

Unlike conventional workflows that often converge on a single line of inquiry, 
AIDA incentivize the insight of long-tail but critical ones. 
By autonomously pairing metrics with a wider array of dimensions, 
AIDA demonstrates a profound potential to navigate complex, real-world industrial data spaces.

\subsection{Environmental Boundary Analysis}
To evaluate whether the model achieves better coupling with the environment, we introduce the concept of environmental boundary analysis.
We conceptualize this data environment as an expansive, interactive map. 
While such data map allows for near-unconstrained exploration, 
the map has physical boundaries defined by the underlying data architecture rules.
An agent may hit a wall through boundary violations, which occur in two primary forms:
Attempting to access metrics or dimensions that do not exist within the defined set;
Requesting combinations of existing metrics and dimensions but which are logically or physically incompatible 
due to data grain or business definitions.

\begin{figure}[htbp]
  \centering
  \includegraphics[width=\linewidth]{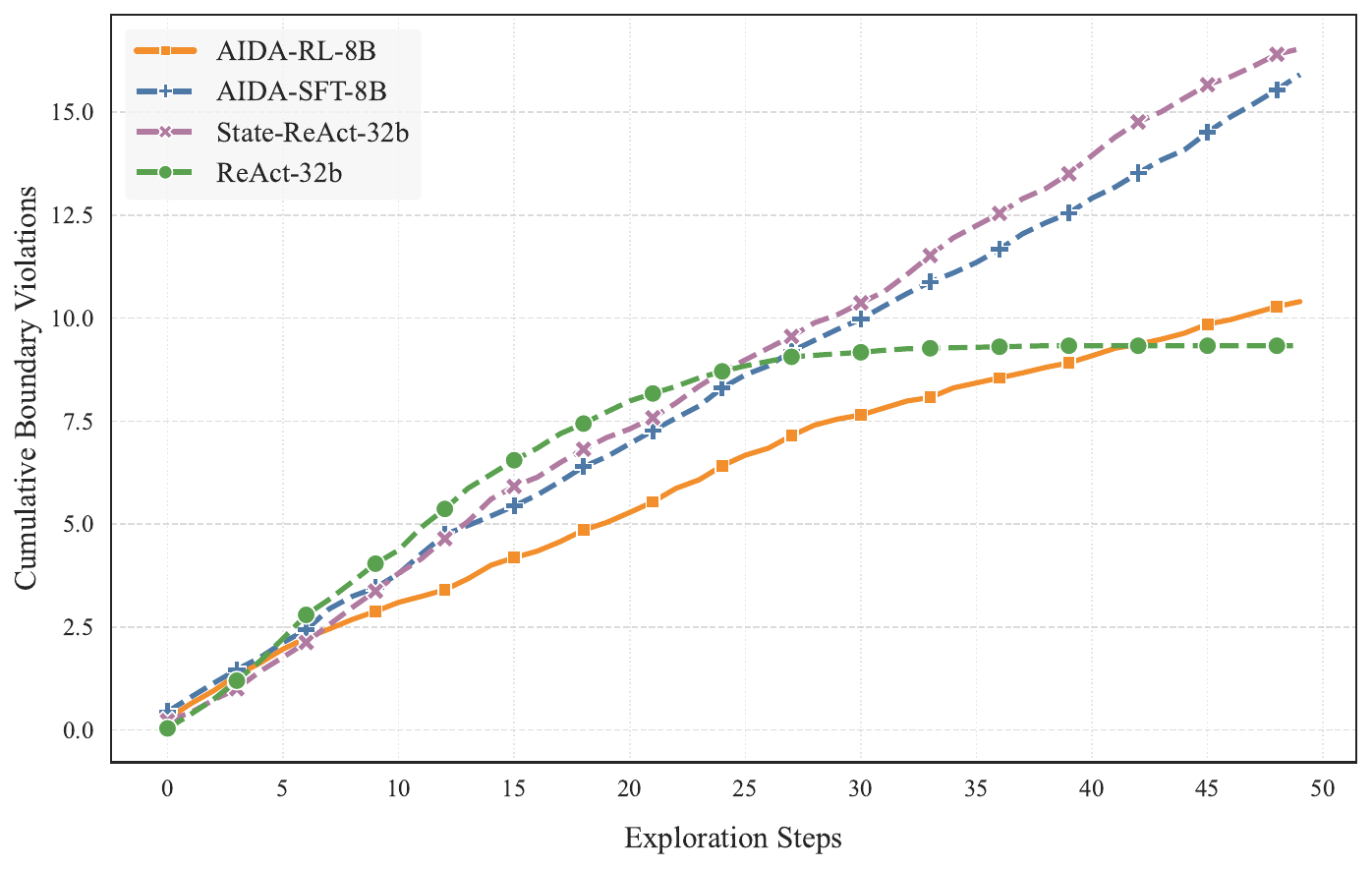} 
  \caption{Environmental boundary analysis.
  The horizontal axis represents the number of exploration steps, 
  and the vertical axis represents the average cumulative number of boundary violations.}
  \label{fig:Perception}
\end{figure}

Figure \ref{fig:Perception} illustrates the cumulative count of these violations during 50-step trajectories. 
The results yield the following insights:
AIDA-RL demonstrates a significant reduction in boundary violations compared to AIDA-SFT and other baselines. 
Crucially, this improvement is an emergent strategic behavior. 
Since the RL objective prioritizes the discovery of valid, high-value insights, 
the agent spontaneously learns to steer away from data that yield no discovery gain. 
Consequently, AIDA-RL maintains a consistently low violation rate even in late-stage reasoning steps, 
whereas baselines often drift toward the environment's boundaries. 
This suggests that our RL framework equips the agent with long-horizon environmental awareness, 
allowing it to select valid metric-dimension combinations even under deep exploration. 
Ultimately, this indicates that the RL process enables the model to internalize the implicit boundaries of business logic, 
achieving a sophisticated balance between aggressive exploration and logical feasibility.

\section{Conclusion}
We propose AIDA, an autonomous framework for data-to-insight discovery within a complex retail environment (200+ metrics) using a proprietary DSL. 
By integrating reinforcement learning with specialized masking and reward strategies, 
AIDA enables stable, high-depth exploration that significantly outperforms existing agents, 
establishing a promising pathway towards truly autonomous business intelligence in industrial-scale systems.

\nocite{langley00}

\bibliography{example_paper}
\bibliographystyle{icml2026}

\newpage
\appendix
\onecolumn

\section{Data development}
\label{appendix:data_env}

The agent environment is a dynamic substrate specifically engineered for high-volume data mining and autonomous retrieval. This section details the environment's architecture, strategic advantages in solving industrial data pain points, and the multi-tier framework that ensures semantic consistency and exploration efficiency.

\subsection{System Architecture}
In industrial scenarios, data retrieval often suffers from inconsistency, semantic drift, and non-idempotent results. The environment addresses these issues through a robust three-tier architecture.

\textbf{Semantic Layer}.
This layer serves as a comprehensive repository, governing standard semantics of metadata for over 200 metrics and 100 dimensions, alongside their respective enumerated values. 

\textbf{Routing Layers}.
The routing layer implements the mapping from semantics to physical tables. Metrics are mounted on \textbf{Detail Tables (DWS)} for granular insight and \textbf{Aggregate Tables (ADS)} for high-frequency queries. The engine intelligently routes queries to minimize response time (RT).

\begin{table}[!ht]
\centering

\label{tab:physical_tables}
\begin{small}
  \caption{Physical Layer Table Schema and Data Scale}
\begin{tabularx}{\textwidth}{l l l X}
\toprule
\textbf{Table Name} & \textbf{Category} & \textbf{Description} \\ \midrule
\texttt{dws\_trd} & DWS & Granular records of trades and order lifecycles. \\
\texttt{dws\_log} & DWS & High-throughput clickstream and behavior logs. \\
\texttt{dim\_usr} & DWS & User profiles and demographics. \\
\texttt{dim\_shop} & DWS & Merchant and category metadata. \\ 
\texttt{dim\_date} & DWS & Temporal metadata and holiday flags. \\ \midrule
\texttt{ads\_shop} & ADS & Pre-aggregated merchant performance indicators. \\ 
\texttt{ads\_brand} & ADS & Pre-aggregated brand-level performance indicators. \\ \bottomrule
\end{tabularx}
\end{small}
\end{table}

The ADS layer reduces RT by over 90\% for standard queries. However, for dynamic exploratory queries that fall back to the DWS layer, the system implements a strict 60-second timeout to ensure resource stability.

\begin{table}[!ht]
\centering
\caption{Response Time Performance}
\label{tab:rt_perf}
\begin{small}
\begin{tabular}{l c c r}
\toprule
\textbf{Query Domain} & \textbf{DWS RT} & \textbf{ADS RT} & \textbf{Latency Reduction} \\ \midrule
Trade Only & 4.0s & 0.5s & -87.5\% \\
Log Only & 18.0s & 0.6s & -96.7\% \\
Cross-Dimension Query & 45.0s & N/A (Exploratory) & \textit{Subject to Timeout} \\ \bottomrule
\end{tabular}
\end{small}
\end{table}

\subsection{Integrated Query Execution and Case Study}
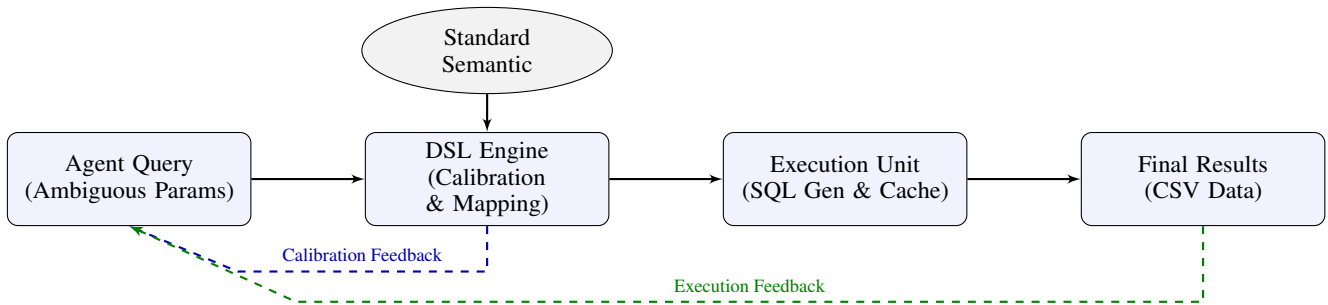
\begin{figure}[!ht]
\centering
\begin{tikzpicture}[
    node distance=1.5cm,
    auto,
    block/.style={rectangle, draw, fill=blue!5, text width=8.5em, text centered, rounded corners, minimum height=3.5em, font=\small},
    cloud/.style={draw, ellipse, fill=gray!10, text width=6em, text centered, minimum height=3em, font=\footnotesize},
    line/.style={draw, -latex', thick}
]
    \node [block] (agent) {Agent Query \\ (Ambiguous Params)};
    \node [block, right=1.5cm of agent] (engine) {DSL Engine \\ (Calibration \& Mapping)};
    \node [block, right=1.5cm of engine] (exec) {Execution Unit \\ (SQL Gen \& Cache)};
    \node [block, right=1.5cm of exec] (output) {Final Results \\ (CSV Data)};
    \node [cloud, above=0.5cm of engine] (semantic) {Standard \\ Semantic};
    \path [line] (agent) -- (engine);
    \path [line] (engine) -- (exec);
    \path [line] (exec) -- (output);
    \path [line] (semantic) -- (engine);
    \path [draw, -latex', dashed, thick, color=blue!70!black] (engine.south) -- ++(0,-0.6) -- node [above, font=\scriptsize] {Calibration Feedback} ++(-3.3,0) -- (agent.south);
    \path [draw, -latex', dashed, thick, color=green!50!black] (output.south) -- ++(0,-1) -- node [above, font=\scriptsize] {Execution Feedback} ++(-12,0) -- (agent.south);
\end{tikzpicture}
\caption{The environment workflow providing two feedback channels: (1) \textbf{Semantic Calibration} and (2) \textbf{Data Evidence}.}
\label{fig:workflow}
\end{figure}

The environment employs a dual-channel feedback mechanism to ensure semantic grounding and evidence-based reasoning as illustrated in Figure~\ref{fig:workflow}.

Agent Query. This entry point is where the agent formulates its analytical intent. 
Given the massive exploration space, initial queries may contain ambiguous parameters that require further 
refinement within the DSL framework.

DSL Engine. The module leverages the Standard Semantic layer to map ambiguous inputs to precise metric and dimension definitions. 
It provides an immediate Calibration Feedback loop, allowing the agent to align its queries with the system's structural logic before execution.
Furthermore, in cases where sorting criteria are unspecified, 
the engine automatically defaults to dimension- and metric-based ordering to ensure the most salient data are prioritized.

Execution Unit. Once calibrated, the DSL is transformed into optimized SQL for data retrieval. 
This unit integrates a caching mechanism to minimize latency, 
ensuring high-fidelity performance even during large-scale iterative mining.

Final Results. The raw output (CSV format) represents the factual basis of the query. 
These results are funneled back to the agent via an Execution Feedback channel, 
providing the empirical evidence necessary for the agent to validate its hypotheses and continue cumulative reasoning.

By decoupling semantic calibration from data execution, this architecture empowers agents to self-correct their analytical paths through two distinct feedback channels, significantly enhancing the efficiency of the "data-to-insights" pipeline.

\begin{table}[!ht]
\centering
\caption{Parameter Calibration and Mapping Logic}
\label{tab:mapping}
\begin{small}
\begin{tabularx}{\textwidth}{l l l l X}
\toprule
\textbf{Entity} & \textbf{Agent Input} & \textbf{Standard Semantic} & \textbf{Status} & \textbf{Engine Correction} \\ \midrule
Dimension & \texttt{week} & \texttt{isWeek} & \textbf{Corrected} & Maps to weekend/weekday flag. \\
Metric & \texttt{aov} & \texttt{netAov} & \textbf{Corrected} & Aligns with Net Order Value formula. \\
Metric & \texttt{gmv} & \texttt{netGMV} & \textbf{Corrected} & Maps to standardized Net GMV. \\
Filter & \texttt{shop="XX"} & \texttt{shopName="XX"} & \textbf{Corrected} & Aligns name to identifier. \\ \bottomrule
\end{tabularx}
\end{small}
\end{table}

\begin{lstlisting}[language=SQL, caption={Synthesized SQL based on Calibrated DSL}]
SELECT
    a.shop_name,
    SUM(net_gmv) AS net_gmv,
    SUM(net_gmv) / COUNT(DISTINCT user_id) AS net_aov
FROM dws_xx_trd_1d a 
JOIN dim_xx_date b ON a.ds = b.ds 
WHERE a.shop_name = 'XX'
GROUP BY b.is_week;
\end{lstlisting}

\begin{minipage}{\linewidth}
\begin{lstlisting}[language=json, caption={Comprehensive Feedback Package (Calibration + Results)}]
{
  "calibration_report": {
    "corrected_dsl": {"metric": ["netAov", "netGMV"], "dimension": ["isWeek"]},
    "notices": "Input 'week' aligned to 'isWeek' (temporal dimension)."
  },
  "execution_results": {
    "data_path": "./xx/2026.csv",
    "preview": [
      {"isWeek": "Weekend", "netAov": 15.2, "netGMV": 45000},
      {"isWeek": "Weekday", "netAov": 12.3, "netGMV": 53632}
    ]
  },
  "status": "Success"
}
\end{lstlisting}
\end{minipage}

\clearpage
\section{Dsl2data Protocol}
\label{appendix:dsl_protocol}

The Dsl2data protocol establishes a standardized interface that facilitates interaction between the agent and the underlying data repository. 
By ensuring semantic consistency, the protocol guarantees that all data retrieval actions are fully executable within the AIDA environment. 
Table~\ref{tab:dsl_schema} details the structural specifications of the Dsl2data protocol. To illustrate its application, a valid request for retrieving data on top-performing merchants is provided.

\begin{table*}[h]
  \caption{Dsl2data Protocol Schema Definitions}
\label{tab:dsl_schema}
\begin{tabularx}{\textwidth}{@{}l l c X@{}}
\toprule
\textbf{Property} & \textbf{Type} & \textbf{Required} & \textbf{Description} \\ \midrule
\texttt{metric} & Array & Yes & List of target metric names (English). \\
\texttt{ds} & Array & Yes & Date range \texttt{["YYYYMMDD", "YYYYMMDD"]} (inclusive). \\
\texttt{dimension} & Array & No & List of dimensions for aggregation. \\
\texttt{filter} & Object & No & Logic conditions (\texttt{and/or}) and \texttt{conditions} array. \\
\texttt{orderBy} & Array & No & Sorting rules containing \texttt{columnEName} and \texttt{orderType}. \\
\texttt{limit} & Integer & No & Maximum rows returned (default: 100). \\
\texttt{compare} & Array & No & Period-over-period analysis (e.g., \texttt{wow}, \texttt{yoy}). \\
\texttt{save\_data\_path} & String & No & Relative path to save the resulting CSV file. \\ \bottomrule
\end{tabularx}
\end{table*}

\begin{lstlisting}[caption={Example request for Dsl2data.}]
{
  "metric": ["Net_GMV"],
  "dimension": ["Gender"],
  "filter": {
    "relation": "and",
    "conditions": [
      {
        "columnEName": "brand_id",
        "queryRule": "in",
        "params": ["xxx"]
      }
    ]
  },
  "ds": ["20251010", "20251110"],
  "orderBy": [
    {"columnEName": "Net_GMV", "orderType": "desc"}
  ],
  "limit": 10
}
\end{lstlisting}

\clearpage
\section{Discovery Gain Reward Judge Prompt}

\begin{lstlisting}[caption={Discovery Gain Reward Judge Prompt},basicstyle=\fontsize{7.5pt}{9pt}\selectfont\ttfamily]
# Background
The agent have provided insights based on data queries and analysis regarding a target problem. You need to assess whether these insights are valid by applying the Pareto Principle.

# Definition of "Key Sub-insight"
A **Key Sub-insight** is an intermediate finding derived through reasoning, induction, or hypothesis testing based on the current data and context. Its characteristics include:
1. Providing substantive progress toward the overall task.
2. Acting as a bridge in the logical chain (connecting previous and subsequent steps).
3. Serving as a core cognitive unit that drives the direction of further exploration.
4. Not yet constituting a final decision or a complete insight.

# Core Principles (Pareto Principle / 80-20 Rule)
- 80% of results are driven by 20% of key factors. A valid key sub-insight must:
  [YES] Focus on key factors (identifying the core of the problem).
  [YES] Ignore the trivial many (avoiding interference from the 80% secondary factors).
  [YES] Reflect the dominant influence of the primary contradiction.
  [YES] Have clear logical support.
- **Invalid insight Scenarios**:
  [NO] The insight is vague, logically inconsistent, or fails to focus on key factors.
  [NO] The insight relies on missing data (e.g., "data not obtained," "no data yet"), making the Pareto Principle inapplicable.
  [NO] The insight is redundant with previous sub-insights.

# Task
**Strictly evaluate the validity of the sub-insight**:
1. Check if the sub-insight aligns with the "Definition of Key Sub-insight."
2. Check if the sub-insight adheres to the Pareto Principle.
3. Output MUST be only: Valid (fully compliant) / Invalid (violates any rule).

# Input Information
- Target Question: {question}
- Previous Key Sub-insight: {insight}
- Current Action: {action}
- Current Environment Observation: {obs}
- Updated Sub-insight (to be evaluated): {insight}

# Response Format
Thought: [Write your analysis and judgment logic here. Keep it logically clear and concise.]
Final Answer: [Output only "Valid" or "Invalid"]
\end{lstlisting}


\end{document}